\documentclass[10pt,twocolumn,letterpaper]{article}

\usepackage{cvpr}
\usepackage{times}
\usepackage{epsfig}
\usepackage{graphicx}
\usepackage{amsmath}
\usepackage{amssymb}

\usepackage[breaklinks=true,bookmarks=false]{hyperref}

\cvprfinalcopy 

\def\cvprPaperID{****} 

\renewcommand{\vec}[1]{\mathbf{#1}}
\newcommand{\matr}[1]{\mathbf{#1}}
\begin{document}

\title{Aggregating Frame-level Features for Large-Scale Video classification}

\author{Shaoxiang Chen$^1$, Xi Wang$^1$, Yongyi Tang$^2$, Xinpeng Chen$^3$, Zuxuan Wu$^1$, Yu-Gang Jiang$^1$\\
$^1$Fudan University~~~~~$^2$Sun Yat-Sen University~~~~~$^3$Wuhan University\\
{\tt\small \{sxchen13,xwang10,zxwu,ygj\}@fudan.edu.cn, tangyy8@mail2.sysu.edu.cn, xinpeng\_chen@whu.edu.cn}
}

\maketitle

\begin{abstract}
   This paper introduces the system we developed for the \emph{Google Cloud \& YouTube-8M Video Understanding Challenge}, which can be considered as a 
   multi-label classification problem defined on top of the large scale YouTube-8M Dataset~\cite{abu2016youtube}. 
   We employ a large set of techniques to aggregate the provided frame-level feature representations and generate video-level predictions,
   including several variants of recurrent neural networks (RNN) and generalized VLAD. 
   We also adopt several fusion strategies to explore the complementarity
   among the models. In terms of the official metric GAP@20 (global average precision at 20), our best fusion model attains 0.84198 on the public 
   50\% of test data and 0.84193 on the private 50\% of test data, ranking 4th out of 650 teams worldwide in the competition.
   
\end{abstract}

\section{Introduction}
In the past several years, we have witnessed the success of Convolutional Neural Networks (CNN) in image understanding tasks like 
 classification \cite{simonyan2014very, szegedy2015going},  segmentation \cite{long2015fully},  
and object detection/localization \cite{ren2015faster}.
It is now well-known that
state-of-the-art CNNs \cite{simonyan2014very,szegedy2016rethinking, szegedy2015going,he2016deep,xie2016aggregated} are superior to the traditional approaches using hand-crafted features. Encouraged by these progresses, many researchers have applied CNNs to video understanding tasks. However, different from images, videos contain not only visual information but also
auditory soundtracks. Also, the continuous frames in videos carry rich motion and temporal information that can hardly be captured by CNN prediction on individual frames, since the CNNs do not naturally handle sequential inputs. In contrast,  Recurrent Neural Networks (RNN) are designed to perform sequential modeling tasks. Combined together, CNN and RNN 
have been proven effective for video analysis \cite{venugopalan2015sequence, wu2016multi}.

Successful models cannot be trained without large-scale annotated datasets like the ImageNet \cite{deng2009imagenet}, FCVID~\cite{jiang2017exploiting} and ActivityNet \cite{caba2015activitynet}.
More recently, YouTube-8M \cite{abu2016youtube} has been released as a benchmark dataset for large-scale video understanding, which contains 8-million videos annotated with over 4,000 class labels and each video is provided with a sequence of frame level features. 
The \emph{Google Cloud \& YouTube-8M Video Understanding Challenge} is based on this new dataset.

Since the challenge only provides pre-computed visual features (using CNN of 
\cite{szegedy2016rethinking}) and audio features without giving the original videos, we can neither 
obtain extra features like optical flows nor compute frame features with different CNNs. Therefore, aggregating the sequence of frame-level features for video-level classification becomes one of the key directions to tackle this challenge.
In our solution, we explore standard RNNs and several variants as means of learning a global descriptor 
from frame level features. We also adopt the idea of a trainable VLAD layer~\cite{arandjelovic2016netvlad} to 
perform feature aggregation in temporal dimension. 
Furthermore, we employ feature transformation to train our models on features from different time
scales. We show that these methods are complementary to 
each other and combining them can produce very competitive results. 

\begin{figure*}
\begin{center}
\includegraphics[width=1.\linewidth]{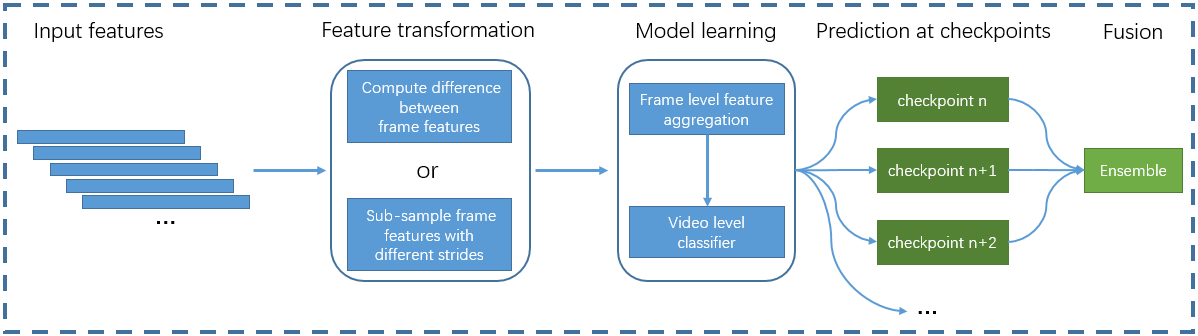}
\end{center}
   \caption{Typical pipeline of a frame-level prediction model. The inputs are a sequence of frame features. During training we save checkpoints for our models
   at different number of iterations, and generate one set of predictions based on each checkpoint's model parameters during testing. These predictions 
   are then fused as the final prediction of this model. Fusing predictions at different iterations is found helpful.}
\label{fig:short}
\end{figure*}

\section{Related Works}

In this section, we briefly review related methods for video classification, particularly those related to our approach developed for the challenge.
In general, the first step is to process video frames or optical flows~\cite{simonyan2014two} by CNNs to get intermediate layer
activations as frame features.
After that, the frame representations are aggregated for video-level prediction. 


In \cite{abu2016youtube,yue2015beyond,wu2016multi}, the authors utilized LSTMs to aggregate frame features extracted by CNNs. Over the years, researchers have developed several solutions to improve the performance 
of RNNs. 
The Gated Recurrent Unit (GRU)~\cite{cho2014learning} can 
often be used in replace of the LSTMs while being more computationally efficient.
Semeniuta \etal \cite{semeniuta2016recurrent} proposed recurrent dropout 
to regularize RNNs during training. The recently proposed Layer Normalization~\cite{ba2016layer} and 
Recurrent Weighted Average (RWA) \cite{ostmeyer2017machine} can help RNNs to converge faster.
In addition, Wu \etal \cite{wu2016google} found residual connections can help train deeply stacked RNNs.

Methods other than RNNs can also be applied for aggregating frame features. 
In \cite{yue2015beyond}, the authors evaluated several feature pooling strategies to 
pool frame features over time. Karpathy \etal \cite{karpathy2014large} used several fusion 
methods to fuse information in the temporal domain. In \cite{arandjelovic2016netvlad}, the authors proposed a new 
generalized VLAD layer to aggregate image representations from CNNs spatially in a trainable manner. Similar idea can be adopted to aggregate a sequence of frame representations temporally.

To further improve the classification performance, fusing multiple models is crucial.
Simonyan \etal \cite{simonyan2014two} used simple linear weighted fusion.
Xu \etal \cite{xu2013feature} proposed a decision-level fusion approach which 
optimizes the weights and thresholds for multiple features in the confidence scores.
Besides fusion at decision level, Jiang \etal \cite{jiang2017exploiting} proposed a feature fusion layer to identify and utilize
the feature correlations in neural networks.

\section{Our Approach}
For video-level models based on averaged frame features, we directly adopt Mixture of Experts (MoE) from \cite{abu2016youtube}.
We use different number of mixtures such as 4, 8 and 16. In the following we mainly focus on frame-level models, which is more important in our approach.

\subsection{Frame-Level Models}
We treat frame-level models as means of aggregating frame features, 
which produce a compact video-level representation. 
By default, we feed this video-level representation into an MoE model as the final classifier. 
Figure~\ref{fig:short} gives a general pipeline of our frame-level model. 

\subsubsection{Variants of RNNs}
Table 1 contains a list of the RNNs we adopted, which are mainly LSTMs, GRUs and their variants.  

All of the RNNs and their variants share the same underlying mechanism.
Generally an RNN cell takes a sequence \((\vec{x}_1, \vec{x}_2, ..., \vec{x}_T)\) as input. It operates on this sequence 
step by step, from \(t=1\) to \(t=T\). At time step \(t\), the RNN cell processes current input \(\vec{x}_t\) and the previous cell state \(\vec{c}_{t-1}\), producing an output \(\vec{h}_t\) at cell state \(\vec{c}_t\). Thus the RNN cell 
can be viewed as a function \(f\) as 
\[\vec{h}_t, \vec{c}_t = f(\vec{x}_t, \vec{c}_{t-1}).\]
After the entire sequence is processed by an RNN cell, we have a sequence of states \((\vec{c}_1, \vec{c}_2, ..., \vec{c}_T)\). Generally we choose \(\vec{c}_T\) to be the representation for this sequence of data.
We can also stack multiple RNN cells. The higher layer RNN cell's input is the output of the lower layer cell. The final state is the concatenation 
of the final states produced by all the layers. Residual connections
can then be added between layers as in~\cite{wu2016google}.

The recurrent dropout variant applies dropout to the cell inputs and outputs at each time step. 
The bidirectional variant has two RNN cells operating on the normal inputs and the reverse \((\vec{x}_T, \vec{x}_{T-1}, ..., \vec{x}_1)\), 
and then concatenates the outputs of the two cells as the final output.

In addition, based on the idea that in many videos the most informative contents appear around the middle of the video, 
we introduce a variant of bidirectional RNN. We split the inputs into two equal subsequences \((\vec{x}_1, \vec{x}_2, ..., \vec{x}_{T/2 - 1})\) and \((\vec{x}_T, \vec{x}_{T-1}, ..., \vec{x}_{T/2})\), feed them to a bidirectional RNN, and then concatenate the final states \(\vec{c}_{T/2-1}\) and \(\vec{c}_{T/2}\) as the video representation. This variant brings some improvement in our final result. 

To better leverage the temporal information captured by the provided frame-level features, we use a simple \emph{feature transformation} method in some of the model training processes.
By taking the difference between adjacent frame pairs as model input, 
the model can make predictions explicitly based on the trend of feature changes. 
Although this feature transformation can cause a performance drop for a single model,
we notice some performance gain by fusing the result with that of the models without using this feature transformation.

Besides, we train our RNNs with frame features at different time scales.
We achieve this by slicing the sequence of frame features into subsequences with equal length in temporal dimension, 
and imposing mean pooling of the frame features in every subsequence to form a sub-sampled sequence as the RNN input. 
The length of the subsequence varies.

Batch normalization \cite{ioffe2015batch} can be applied to the RNN cell output to accelerate convergence.

\begin{table}
\begin{center}
\begin{tabular}{|l|c|}
\hline
Model & Variations \\
\hline\hline
LSTM & --- \\
LSTM & Layer normalization \& Recurrent dropout \\
RNN & Residual connections \\
GRU & --- \\
GRU & Bi-directional\\
GRU & Recurrent dropout\\
GRU & Feature transformation \\
RWA & --- \\
NetVLAD & --- \\
DBoF & --- \\
\hline
\end{tabular}
\end{center}
\caption{Frame level models. No variation in the second column means the original implementation is adopted. DBoF is from~\cite{abu2016youtube}.}
\end{table}

\subsubsection{VLAD Aggregation}

In~\cite{arandjelovic2016netvlad}, Arandjelovic \etal proposed NetVLAD, using a VLAD layer to pool descriptors extracted from CNNs. 
The pooling operates on the spatial dimensions of the descriptors. Here we borrow this idea to pool 
video frame features temporally. Given the sequence of frame features \((\vec{x}_1, \vec{x}_2, ..., \vec{x}_T)\), which is \(T\times D\)-dimensional and \(T\) may 
vary across samples since the video length is different. We wish to pool frame features into a fixed length \(K\times D\)-dimensional descriptor. 
Here \(K\) is a parameter we can adjust as a trade-off between computation cost and performance.

We first randomly sample \(S\) out of \(T\) frame features, denoted by \(\matr{R}=(\vec{r}_1, \vec{r}_2, ..., \vec{r}_S)\). \(\matr{R}\) can be viewed as a \(S\times D\)-dimensional matrix. The cluster mean is denoted by \((\vec{u}_1, \vec{u}_2, ..., \vec{u}_K)\), which is a \(K\times D\)-dimensional trainable parameter.
We compute the strength of association by 1D-convolving \(\matr{R}\) into a \(S\times K\)-dimensional matrix \(\matr{A}=(\vec{a}_1, \vec{a}_2, ..., \vec{a}_S)\). We use a 1-D convolution kernel with width of 1 and output channel of \(k\) here.
Then we apply soft-max to \(\matr{A}\), so that \(\sum_{k=1}^{K} a_{ik} = 1\). The aggregated descriptor is computed by 
\[\vec{v}_k = \sum_{i=1}^{S} a_{ik}(\vec{r}_i-\vec{u}_k)\]

The resulted descriptors \((\vec{v}_1, \vec{v}_2, ..., \vec{v}_K)\) are concatenated to be the new video level representation. 
Since this aggregation method is different 
from the RNN based methods, the results produced by this model can be a good complement to that of the RNNs during fusion. Compared with RNNs, the computational cost of this method is lower.

\subsection{Label Filtering}
The class distribution in the YouTube-8M dataset is imbalanced. Some classes have many positive samples, while some have much fewer samples. 
In order to better predict those labels with relatively small occurrence probability, we use label filters in some of the model training processes. 
The labels with high occurrence probability are discarded during training since they are well-trained for other models based on the full set of the labels. 
Two filter thresholds are used in our approach, making these models focusing only on 2,534 and 3,571 classes with fewer positive training samples, respectively.

\subsection{Model Fusion Strategies}

Our final prediction is a linear weighted fusion of prediction scores produced by multiple models. 
Specifically, the fusion is done in two stages.

\textbf{Stage 1}. We get predictions from multiple model checkpoints saved during training. The checkpoint is chosen 
after the model is trained for more than 3 epochs. We fuse these predictions as a final result for this model.
This stage can be regarded as intra-model fusion.

\textbf{Stage 2}. We fuse predictions from different models generated in Stage 1 to get our final prediction.
This can be regarded as inter-model fusion.

We try the following three simple strategies to determine the fusion weights:

\emph{Empirical fusion weights:} Weights are assigned based on empirical experience of model performance. Better models are assigned with higher weights.

\emph{Brute-force Search of fusion weights:} 
On the validation set, we can perform grid-search of fusion weights to identify the best model combination.

\emph{Learning for fusion weights:} 
We can also train a linear regression model to learn the fusion weights on the validation set.

\subsection{Implementation Details}

All of our models are trained based on the starter TensorFlow code\footnote{https://github.com/google/youtube-8m}.
Layer and Batch normalization are directly available in TensorFlow.
For RWA, we take the authors' open source implementation\footnote{https://github.com/jostmey/rwa}. 
For residual connections in RNN, we use an open source implementation\footnote{https://github.com/NickShahML/tensorflow\_with\_latest\_papers}.


We concatenate the provided visual and audio features before model training.  
For our NetVLAD model, we separately process visual and audio features through VLAD layer, 
and then concatenate them afterwards. We generally stack 2 layers of RNNs. 


Please refer to this link\footnote{http://github.com/forwchen/yt8m} for more details.



\section{Evaluation}
The models are trained and evaluated on machines with the following settings: OS of Ubuntu 14.04 with gcc version 
4.8.4, CUDA-8.0, TensorFlow 1.0.0 and GPU of GeForce GTX TITAN X.
The learning rates of our RNN models and NetVLAD models are 0.001 with exponentially decay after each epoch with a decay rate of 0.95. 
The batch sizes are 128 or 256. Model checkpoints are automatically 
saved during training every 0.5 hours.

\subsection{The Challenge Dataset}
The videos in the YouTube-8M dataset are sampled uniformly on YouTube to preserve the diverse distribution of popular contents. Each video is between 120 and 500 seconds long. The selected videos are decoded at 1 frame-per-second up to the first 360 seconds (6 minutes). The decoded frames are fed into the Inception network \cite{szegedy2016rethinking} and the ReLu activation of the last hidden layer is extracted. After applying PCA to reduce feature dimensions to 1024, the resulted features are provided in the challenge as frame-level features. The video-level features are simply the mean of all the frame features of the video. 
There are 4,716 classes in total. A video sample may have multiple labels and the average number of classes per video is 1.8. Table 2 gives the dataset partition used in this challenge competition.

\begin{table}[ht!]
\begin{center}

\begin{tabular}{|c|c|}
\hline
Partition & Number of Samples\\
\hline\hline
Train &4,906,660 \\
Validate & 1,401,828\\
Test & 700,640\\
Total & 7,009,128 \\

\hline
\end{tabular}
\end{center}
\caption{Dataset partition of the YouTube-8M dataset in the challenge.}
\end{table}

\subsection{Evaluation Metric}
In the challenge, the predictions are evaluated by Global Average Precision (GAP) at 20. For a result with 
\(N\) predictions (label/confidence pairs) sorted by its confidence score, the GAP is computed as:
\[GAP=\sum_{i=1}^{N} p(i)\Delta r(i),\] 
where \(N\) is the number of final predictions (if there are 20 predictions for each video, then $N = 20 * \#Videos$ ), $p(i)$ is the precision for the first $i$ predictions, and $\Delta r(i)$ is the change in recall. We denote the total number of positives in these \(N\) predictions as $m$. If prediction $i$ is correct then $\Delta r(i)=1/m$, otherwise $\Delta r(i)=0$.

\subsection{Results}
\begin{table*}[ht!]
\begin{center}
\begin{tabular}{|l|l|l|c|}
\hline
Model & \#Iterations of checkpoints & Intra-model fusion weights & Inter-model fusion weights\\
\hline\hline
LSTM & 353k, 323k, 300k, 280k & 0.4, 0.3, 0.2, 0.1 & 1.0 \\
GRU &  69k, 65k, 60k, 55k & 0.4, 0.3, 0.2, 0.1 & 1.0 \\
RWA & 114k, 87k, 75k, 50k & 0.4, 0.3, 0.2, 0.1 & 1.0 \\
GRU w. recurrent dropout & 56k, 50k, 46k, 40k, 35k & 0.4, 0.3, 0.2, 0.05, 0.05 & 1.0 \\
NetVLAD & 24k, 21k, 19k, 16k, 13k & 0.4, 0.3, 0.2, 0.05, 0.05 & 1.0 \\
MoE & 127k, 115k, 102k, 90k & 0.4, 0.3, 0.2, 0.1 & 0.5 \\
DBoF & 175k, 150k, 137k, 122k, 112k  & 0.4, 0.3, 0.2, 0.05, 0.05 & 0.5 \\
GRU w. batch normalization & 86k, 74k, 65k, 49k & 0.4, 0.3, 0.2, 0.1 & 0.25 \\
Bidirectional GRU & 53k, 45k, 35k & 0.5, 0.3, 0.2 & 0.25 \\
\hline
\end{tabular}
\end{center}
\caption{Details of the models used in our approach. Every model produces 3 to 5 predictions on the test set with parameters obtained on different training checkpoints and these predictions 
are firstly linearly fused with the intra-model weights. The resulted predictions of these models are then fused with the inter-model weights. }
\end{table*}

In the challenge, our submissions are generated by fusing multiple models. Here provide details of one setting with competitive results. Table 3 gives the details. This setting produces a GAP@20 of 0.83996 on the public 50\% of test set.

We further provide results of some models and their fusion in Table 4.
We can see that the performance can always be improved by appropriately fusing more model predictions.
The single NetVLAD is from the prediction of one model trained for 10k iterations. Our NetVLAD produces competitive results to the RNNs, which is appealing considering its low computation cost. 
Ensemble 1 is the fusion of all the models in Table 3. 
Ensemble 2 produces our best result, coming from the fusion of Ensemble 1 with some additional models. Details can be found at this link\footnote{http://github.com/forwchen/yt8m}.

\begin{table}[ht!]
\begin{center}
\caption{Results on the public 50\% of test set. See texts for details.}.
\begin{tabular}{c|c}
\hline
Model &  GAP\\
\hline\hline
Single NetVLAD & 0.79175 \\
\hline
NetVLAD multi-model fusion & 0.80895 \\
LSTM  multi-model fusion & 0.81571 \\
GRU  multi-model fusion & 0.81786 \\
\hline
Ensemble 1 & 0.83996 \\
Ensemble 2 & \textbf{0.84198} \\
\hline
\end{tabular}
\end{center}

\end{table}

\section{Conclusion}
We have introduced an approach to aggregate frame-level features for large-scale video classification. 
We showed that fusing multiple models is always helpful. We also proposed a variant of VLAD to aggregate sequence of frame features temporally, which 
 can produce good results with lower computational cost than RNN. Adding all the carefully designed strategies together, our system ranked 4th out of 650 teams worldwide in the challenge competition. 
 
{\small
\bibliographystyle{ieee}
\bibliography{egbib}

\begin{thebibliography}{10}\itemsep=-1pt

\bibitem{abu2016youtube}
S.~Abu-El-Haija, N.~Kothari, J.~Lee, P.~Natsev, G.~Toderici, B.~Varadarajan,
  and S.~Vijayanarasimhan.
\newblock Youtube-8m: A large-scale video classification benchmark.
\newblock {\em arXiv preprint arXiv:1609.08675}, 2016.

\bibitem{arandjelovic2016netvlad}
R.~Arandjelovic, P.~Gronat, A.~Torii, T.~Pajdla, and J.~Sivic.
\newblock Netvlad: Cnn architecture for weakly supervised place recognition.
\newblock In {\em Proceedings of the IEEE Conference on Computer Vision and
  Pattern Recognition}, pages 5297--5307, 2016.

\bibitem{ba2016layer}
J.~L. Ba, J.~R. Kiros, and G.~E. Hinton.
\newblock Layer normalization.
\newblock {\em arXiv preprint arXiv:1607.06450}, 2016.

\bibitem{caba2015activitynet}
F.~Caba~Heilbron, V.~Escorcia, B.~Ghanem, and J.~Carlos~Niebles.
\newblock Activitynet: A large-scale video benchmark for human activity
  understanding.
\newblock In {\em Proceedings of the IEEE Conference on Computer Vision and
  Pattern Recognition}, pages 961--970, 2015.

\bibitem{cho2014learning}
K.~Cho, B.~Van~Merri{\"e}nboer, C.~Gulcehre, D.~Bahdanau, F.~Bougares,
  H.~Schwenk, and Y.~Bengio.
\newblock Learning phrase representations using rnn encoder-decoder for
  statistical machine translation.
\newblock {\em arXiv preprint arXiv:1406.1078}, 2014.

\bibitem{deng2009imagenet}
J.~Deng, W.~Dong, R.~Socher, L.-J. Li, K.~Li, and L.~Fei-Fei.
\newblock Imagenet: A large-scale hierarchical image database.
\newblock In {\em Computer Vision and Pattern Recognition, 2009. CVPR 2009.
  IEEE Conference on}, pages 248--255. IEEE, 2009.

\bibitem{he2016deep}
K.~He, X.~Zhang, S.~Ren, and J.~Sun.
\newblock Deep residual learning for image recognition.
\newblock In {\em Proceedings of the IEEE Conference on Computer Vision and
  Pattern Recognition}, pages 770--778, 2016.

\bibitem{ioffe2015batch}
S.~Ioffe and C.~Szegedy.
\newblock Batch normalization: Accelerating deep network training by reducing
  internal covariate shift.
\newblock {\em arXiv preprint arXiv:1502.03167}, 2015.

\bibitem{jiang2017exploiting}
Y.-G. Jiang, Z.~Wu, J.~Wang, X.~Xue, and S.-F. Chang.
\newblock Exploiting feature and class relationships in video categorization
  with regularized deep neural networks.
\newblock {\em IEEE Transactions on Pattern Analysis and Machine Intelligence},
  2017.

\bibitem{karpathy2014large}
A.~Karpathy, G.~Toderici, S.~Shetty, T.~Leung, R.~Sukthankar, and L.~Fei-Fei.
\newblock Large-scale video classification with convolutional neural networks.
\newblock In {\em Proceedings of the IEEE conference on Computer Vision and
  Pattern Recognition}, pages 1725--1732, 2014.

\bibitem{long2015fully}
J.~Long, E.~Shelhamer, and T.~Darrell.
\newblock Fully convolutional networks for semantic segmentation.
\newblock In {\em Proceedings of the IEEE Conference on Computer Vision and
  Pattern Recognition}, pages 3431--3440, 2015.

\bibitem{ostmeyer2017machine}
J.~Ostmeyer and L.~Cowell.
\newblock Machine learning on sequential data using a recurrent weighted
  average.
\newblock {\em arXiv preprint arXiv:1703.01253}, 2017.

\bibitem{ren2015faster}
S.~Ren, K.~He, R.~Girshick, and J.~Sun.
\newblock Faster r-cnn: Towards real-time object detection with region proposal
  networks.
\newblock In {\em Advances in neural information processing systems}, pages
  91--99, 2015.

\bibitem{semeniuta2016recurrent}
S.~Semeniuta, A.~Severyn, and E.~Barth.
\newblock Recurrent dropout without memory loss.
\newblock {\em arXiv preprint arXiv:1603.05118}, 2016.

\bibitem{simonyan2014two}
K.~Simonyan and A.~Zisserman.
\newblock Two-stream convolutional networks for action recognition in videos.
\newblock In {\em Advances in neural information processing systems}, pages
  568--576, 2014.

\bibitem{simonyan2014very}
K.~Simonyan and A.~Zisserman.
\newblock Very deep convolutional networks for large-scale image recognition.
\newblock {\em arXiv preprint arXiv:1409.1556}, 2014.

\bibitem{szegedy2015going}
C.~Szegedy, W.~Liu, Y.~Jia, P.~Sermanet, S.~Reed, D.~Anguelov, D.~Erhan,
  V.~Vanhoucke, and A.~Rabinovich.
\newblock Going deeper with convolutions.
\newblock In {\em Proceedings of the IEEE Conference on Computer Vision and
  Pattern Recognition}, pages 1--9, 2015.

\bibitem{szegedy2016rethinking}
C.~Szegedy, V.~Vanhoucke, S.~Ioffe, J.~Shlens, and Z.~Wojna.
\newblock Rethinking the inception architecture for computer vision.
\newblock In {\em Proceedings of the IEEE Conference on Computer Vision and
  Pattern Recognition}, pages 2818--2826, 2016.

\bibitem{venugopalan2015sequence}
S.~Venugopalan, M.~Rohrbach, J.~Donahue, R.~Mooney, T.~Darrell, and K.~Saenko.
\newblock Sequence to sequence-video to text.
\newblock In {\em Proceedings of the IEEE International Conference on Computer
  Vision}, pages 4534--4542, 2015.

\bibitem{wu2016google}
Y.~Wu, M.~Schuster, Z.~Chen, Q.~V. Le, M.~Norouzi, W.~Macherey, M.~Krikun,
  Y.~Cao, Q.~Gao, K.~Macherey, et~al.
\newblock Google's neural machine translation system: Bridging the gap between
  human and machine translation.
\newblock {\em arXiv preprint arXiv:1609.08144}, 2016.

\bibitem{wu2016multi}
Z.~Wu, Y.-G. Jiang, X.~Wang, H.~Ye, and X.~Xue.
\newblock Multi-stream multi-class fusion of deep networks for video
  classification.
\newblock In {\em Proceedings of the 2016 ACM on Multimedia Conference}, pages
  791--800. ACM, 2016.

\bibitem{xie2016aggregated}
S.~Xie, R.~Girshick, P.~Doll{\'a}r, Z.~Tu, and K.~He.
\newblock Aggregated residual transformations for deep neural networks.
\newblock {\em arXiv preprint arXiv:1611.05431}, 2016.

\bibitem{xu2013feature}
Z.~Xu, Y.~Yang, I.~Tsang, N.~Sebe, and A.~G. Hauptmann.
\newblock Feature weighting via optimal thresholding for video analysis.
\newblock In {\em Proceedings of the IEEE International Conference on Computer
  Vision}, pages 3440--3447, 2013.

\bibitem{yue2015beyond}
J.~Yue-Hei~Ng, M.~Hausknecht, S.~Vijayanarasimhan, O.~Vinyals, R.~Monga, and
  G.~Toderici.
\newblock Beyond short snippets: Deep networks for video classification.
\newblock In {\em Proceedings of the IEEE conference on computer vision and
  pattern recognition}, pages 4694--4702, 2015.

\end{thebibliography}
}

\end{document}